\documentclass[letterpaper, 10 pt, conference]{ieeeconf}  

\IEEEoverridecommandlockouts                              

\usepackage{xcolor}
\usepackage{graphicx}
\usepackage{multirow}

\usepackage{amsmath}
\usepackage{comment}
\usepackage{mathabx}
\usepackage{siunitx}
\usepackage{import}

\title{\LARGE \bf
Measuring DNA Microswimmer Locomotion in Complex Flow Environments
}

\author{Taryn Imamura*$^{1}$, Teresa A. Kent*$^{2}$, Rebecca E. Taylor$^{1}$ and Sarah Bergbreiter$^{1}$
\thanks{*These authors contributed equally to this work}
\thanks{This work was supported in part by National Science Foundation (NSF) grants 1739308 and 2132886, an NSF GRFP DGE1745016/DGE2140739, a seed grant from the Manufacturing Futures Institute at Carnegie Mellon University, and Air Force Office of Scientific Research MURI award number FA9550-19-1-0386.}
\thanks{$^{1}$ Mechanical Engineering, Carnegie Mellon University, Pittsburgh, PA, USA Corresponding Author: Taryn Imamura
 (tri@andrew.cmu.edu)}
\thanks{$^{2}$ Robotics Institute, Carnegie Mellon University, Pittsburgh, PA, USA}
}

\begin{document}

\maketitle
\thispagestyle{empty}
\pagestyle{empty}

\begin{abstract}

Microswimmers are sub-millimeter swimming microrobots that show potential as a platform for controllable locomotion in applications including targeted cargo delivery and minimally invasive surgery. To be viable for these target applications, microswimmers will eventually need to be able to navigate in environments with dynamic fluid flows and forces. Experimental studies with microswimmers towards this goal are currently rare because of the difficulty isolating intentional microswimmer motion from environment-induced motion. In this work, we present a method for measuring microswimmer locomotion within a complex flow environment using fiducial microspheres. By tracking the particle motion of ferromagnetic and non-magnetic polystyrene fiducial microspheres, we capture the effect of fluid flow and field gradients on microswimmer trajectories. We then determine the field-driven translation of these microswimmers relative to fluid flow and demonstrate the effectiveness of this method by illustrating the motion of multiple microswimmers through different flows. 

\end{abstract}

\section{Introduction}

Whether they are phytoplankton, bacteria, or spermatozoa, nature is full of biological microswimmers \cite{guasto_fluid_2012, lauga_bacterial_2016, ishimoto_fluid_2015}. 
The advanced locomotion of these microorganisms as they avoid predators and locate food in low Reynolds environments \cite{lauga_bacterial_2016, qiu_navigation_2022} and navigate  complex biological flows \cite{ghosh_slippery_2023, qiu_navigation_2022} is desirable in their synthetic counterparts. Synthetic microswimmers are swimming robots less than \SI{1}{\milli\meter} in size that have demonstrated great potential for applications in targeted cargo delivery, minimally invasive surgery, in vivo diagnostics, and water remediation \cite{alapan_soft_2018, jeon_magnetically_2019, wang_nanomicroscale_2012, soler_catalytic_2014, gao_environmental_2014}. To achieve these desired applications, synthetic microswimmers must be capable of reliably performing desired actions in the presence of external stimuli, such as fluid flow. 

Computational and numerical studies on naturally-occurring microorganisms and microswimmer models have revealed theoretical insights into how we can best design and control synthetic microswimmers that are specialized to various biophysical environments \cite{caldag_steering_2020, muinos-landin_reinforcement_2021, ghosh_slippery_2023, feng_dynamics_2023, qiu_navigation_2022, serrano_purcells_2022}. Some of these insights have been validated through experimental studies that have demonstrated microswimmer locomotion \cite{alvarez_reconfigurable_2021, tierno_controlled_2008, zhang_artificial_2010, du_reconfigurable_2018, harmatz_customizable_2020, kei_cheang_multiple-robot_2014, cheang_self-assembly_2015, maier_magnetic_2016, ullrich_swimming_2016, barbot_controllable_2014, huang_adaptive_2019} and the creation of local flow fields driven by microswimmer undulation and rotation \cite{khaderi_magnetically-actuated_2011, diller_rotating_2011, xie_programmable_2019, zhou_ye_rotating_2014}.
To simplify experimental design, the majority of these studies were conducted in sealed test arenas with negatively buoyant microswimmers, thereby creating idealized test environments with minimized flow or effect of flow on the microswimmers. 

However, studying microswimmers in non-idealized conditions (i.e. suspended in a liquid and subject to external stimuli such as flow) is more challenging, especially for microswimmers that are less than \SI{100}{\micro\meter} in size. It can be difficult to draw conclusions about the causality of microswimmer motion or conduct path planning without a methodology to either create a controlled flow environment \cite{dey_oscillatory_2022, sharan_microfluidics_2021} or measure flow within a given environment. 
While methods such as particle image velocimetry (PIV) and micro PIV ($\mu$PIV) can be used to track and visualize flow \cite{gemmell2014new,zickus20184d,chen2013high}, they are less effective for microswimmers that are smaller than the size of the tracking particles (typically \SI{10}{\micro\meter} - \SI{100}{\micro\meter}). 
These tracking particles could negatively impact experiments by impeding the motion of microswimmers, adhering to the microswimmers, or making microswimmers more difficult to identify and track.

\begin{figure}[t]
\centering
\includegraphics[width=0.85\linewidth]{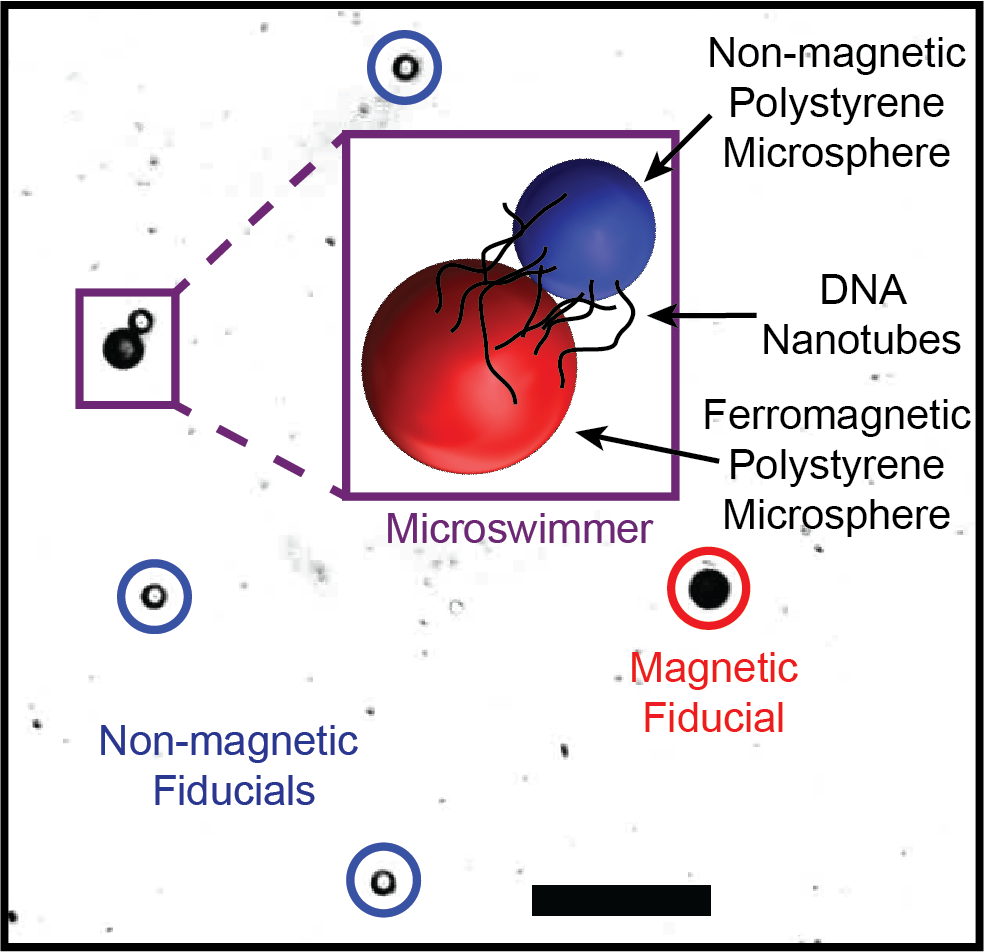}
\caption{Microscope image of non-magnetic fiducial microspheres, a magnetic fiducial microsphere, and a microswimmer. The inset shows the microswimmer structure (a non-magnetic polystyrene microsphere connected to a ferromagnetic polystyrene microsphere by a network of DNA nanotubes). 
The fiducial microspheres allow us to track fluid flow during experiments. Scale bar is \SI{50}{\micro\meter}.}
\label{fig:MSandFid}
\end{figure}
The primary contribution of this work is a method for measuring microswimmer locomotion within a complex flow environment using fiducial microspheres (Fig.~\ref{fig:MSandFid}). We use colloidal microswimmers, which are a class of modular microswimmers that are comprised of specialized micro- or nano-scale components that are linked together \cite{ebbens_self-assembled_2010,ni_hybrid_2017, cui_molecular_2023, lyu_active_2023}. Similar to those in \cite{taryn_imamura_fabrication_2023, harmatz_customizable_2020}, our microswimmers are approximately \SI{17}{\micro\meter} in length and composed of ferromagnetic and non-magnetic polystyrene microspheres that are connected by a flexible linkage of DNA nanotubes (Fig.~\ref{fig:MSandFid}). 
To track the microswimmer movement in solution, we leverage the assumption that flow within a given region can be tracked using particles of similar size and density of our microswimmers. 
We predict that microswimmers will rotate in response to magnetic field inputs, and that oscillating magnetic fields will result in net translation for microswimmers, but not for fiducial microspheres \cite{benjaminson_buoyant_2023, singh_grover_motion_2019}. 
As a secondary contribution, we use this methodology to demonstrate multiple, different suspended microswimmers locomoting in complex flow environments. To the best of our knowledge this work is the first experimental study in which the locomotion of suspended colloidal microswimmers is quantified within the presence of flow, and provides a process through which the ground truth movement of colloidal microswimmers can be obtained in the presence of complex fluid flow.

\section{Methods}
\begin{figure}[t]
\centering
\includegraphics[width=8 cm]{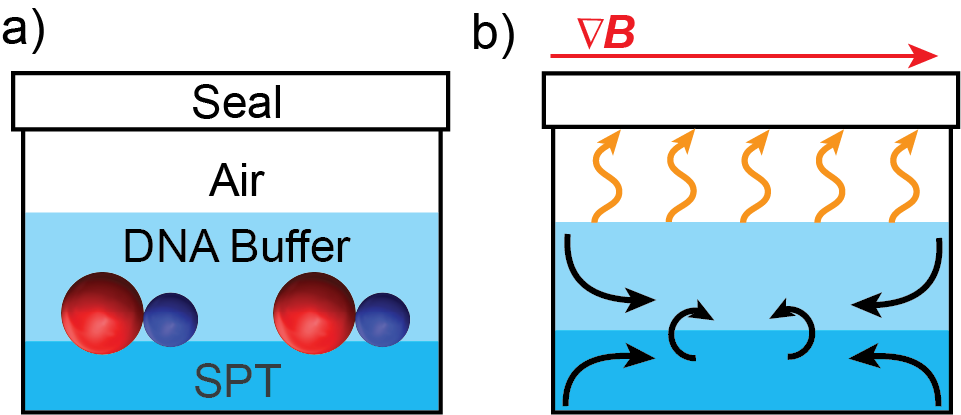}
\caption{a) Chamber and fluid contents used in microswimmer swimming studies. b) Possible sources of fluid flow and microswimmer drift include evaporation (orange arrows), fluid mixing between solution layers (black arrows), and magnetic field gradients (red arrow).
}
\label{fig:FluidWeather}
\end{figure}
\subsection{Microswimmer Manufacturing}

We chose pairs of large ferromagnetic and small non-magnetic microspheres connected by DNA linkages for our microswimmer design (Fig.~\ref{fig:MSandFid}) because this design will allow the microswimmers to break the ‘Scallop Theorem.’ 
This theorem states that 
microswimmers require non-reciprocal body deformation or asymmetry to achieve locomotion in low Reynolds number environments \cite{purcell_life_1977, du_reconfigurable_2018}. 
The asymmetric design of the microswimmers and the inclusion of compliant DNA linkages provide the non-reciprocal motion needed for this structure to achieve net displacement in the low Reynolds regimes \cite{benjaminson_buoyant_2023, singh_grover_motion_2019}.

Following procedures from Imamura et al. \cite{taryn_imamura_fabrication_2023}, streptavidin-coated ferromagnetic and non-magnetic polystyrene microspheres (\SI[separate-uncertainty = true,multi-part-units=single]{10.3(8)}{\micro\meter} and \SI[separate-uncertainty = true,multi-part-units=single]{6.8(5)}{\micro\meter} respectively, Spherotech Inc.) were assembled via Templated Assembly by Selective Removal (TASR). 
Microswimmers were removed from the template using a layer of dehydrated Tris-acetate-EDTA buffer (1xTAE) with MgCl\textsubscript{2} (12.5 mM) that was then dissolved to release the microswimmers into solution. Unbound ferromagnetic and non-magnetic microspheres were also deposited into solution and used as fiducials to track fluid flow.

Because the goal of this study was to evaluate microswimmer motion when subject to magnetic fields and flow, it was important that microswimmers and fiducial microspheres remained suspended in solution. The density of the DNA buffer alone was too low to keep the microspheres suspended, so an additional solution of sodium polytungstate (SPT, 400 mM, 71913, Sigma Aldrich) was added \cite{smyrlaki_solid_2023}. Microswimmers and fiducial microspheres were deposited into PET petri dishes (430165, 33mm, Corning Inc.) which contained a 2:1 solution of SPT and 1xTAE with 12.5 MgCl\textsubscript{2}. The SPT is denser than 1xTAE, and formed a fluid layer to keep the microswimmers suspended throughout the study (Fig. \ref{fig:FluidWeather}a).

All materials were brought to room temperature prior to assembly to minimize thermal effects and the system was left for one hour to achieve steady-state and minimize fluid mixing. Despite these measures, substantial fluid circulation, particle movement, and condensation were observed in the system. Based on these observations, we speculate that this flow resulted from evaporation of the test solution and continued fluid mixing between the SPT and 1xTAE layers. Dye studies showed that the SPT and 1xTAE layers fully separate over 8-12 hours, but this rest period was deemed unrealistic because it exceeded the period in which the DNA nanotubes were most mechanically stable, and the layers could be easily remixed by slight disturbances. We also speculate that the motion of microswimmers and magnetic fiducials may be affected by magnetic field gradients. These magnetic gradients may arise due to nonidealities in the Helmholtz coil test setup including temperature effects, coil misplacement, and poorly regulated control current  (Fig.~\ref{fig:FluidWeather}b).

\subsection{Microswimmer Actuation}

\begin{figure}[t]
\centering
\includegraphics[width=0.8\linewidth]{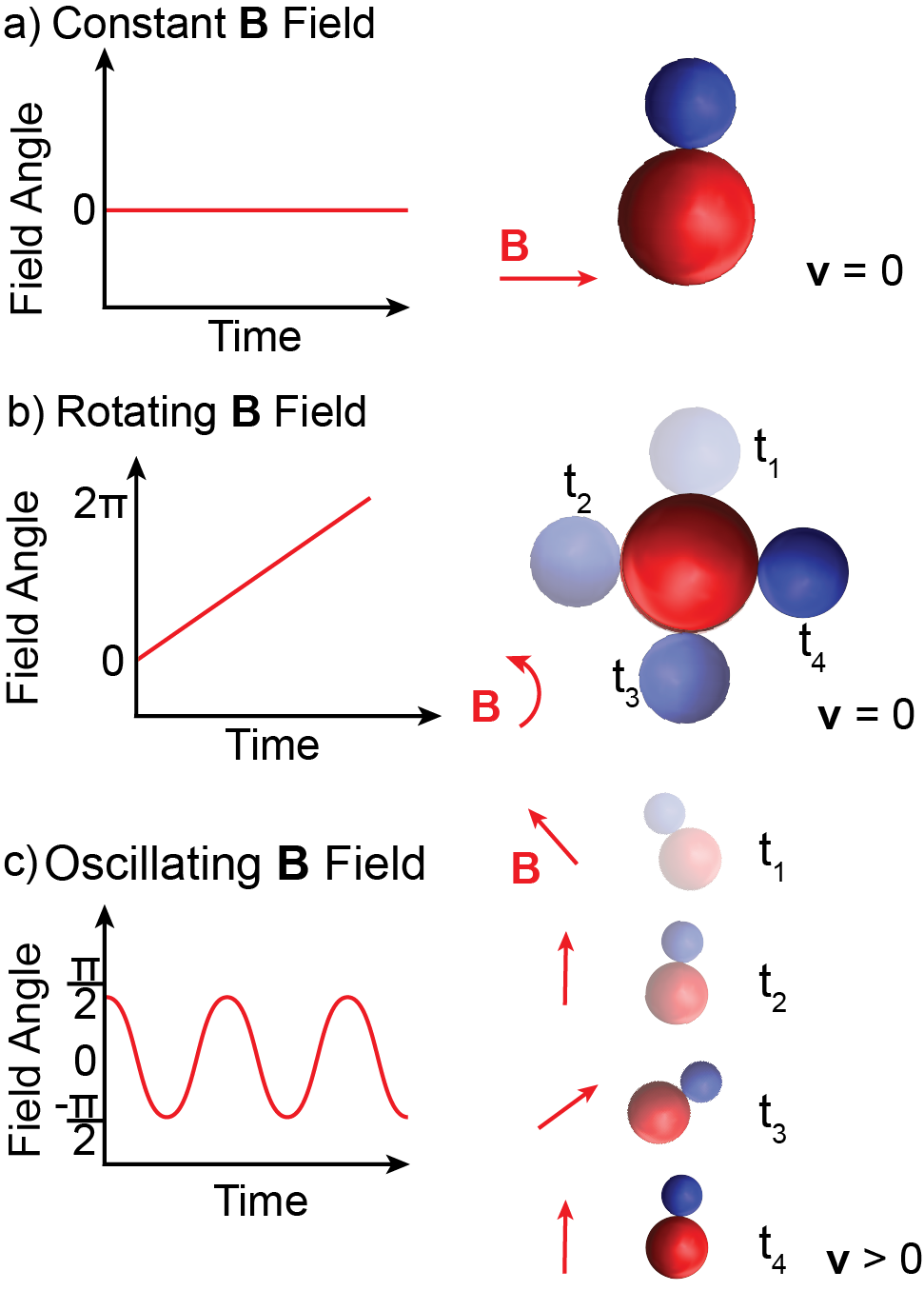}
\caption{Diagram showing microswimmer motion due to a) constant, b) rotating, and c) oscillating magnetic fields. Only the oscillating field generates the non-reciprocal body deformations needed to break the ``Scallop Theorem" and achieve net translation.}
\label{fig:FieldSchema}
\end{figure}

\begin{figure}[t]
\centering
\includegraphics[width=0.75\linewidth]{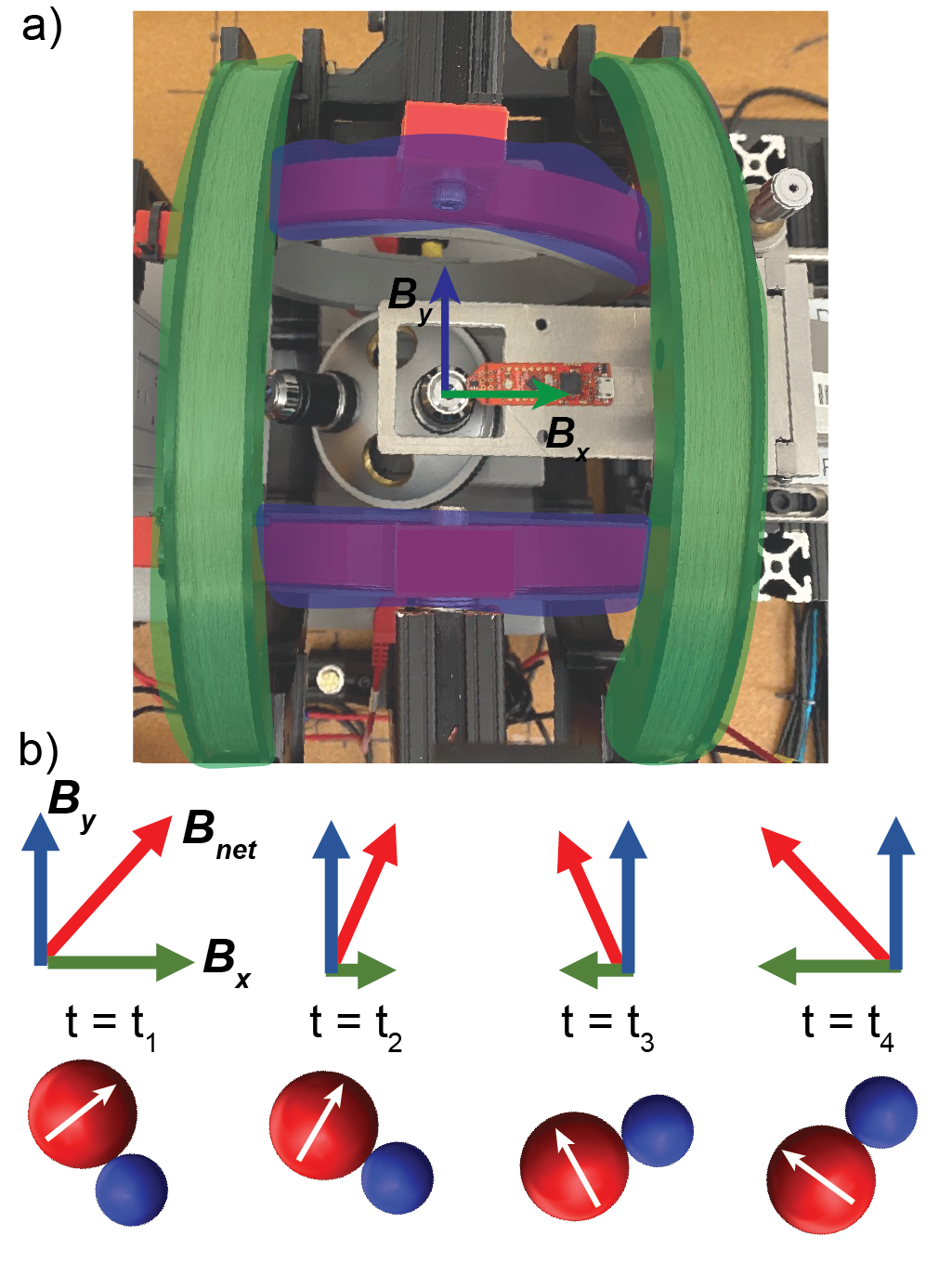}
\caption{a) Orthogonal pairs of Helmholtz coils were used to generate uniform, controlled magnetic field inputs. b) Each coil (green: Bx, blue By) generated a component vector of the net magnetic field (red).}
\label{fig:Coils}
\end{figure}

With the goal of determining how different magnetic field inputs affect microswimmers in the presence of flow, we subjected non-magnetic fiducial particles, magnetic fiducial particles,  and microswimmers to constant, rotating, and oscillating magnetic fields inputs (Fig. \ref{fig:FieldSchema}a-c). As done in \cite{benjaminson_buoyant_2023, taryn_imamura_fabrication_2023, harmatz_customizable_2020}, these fields were generated by a pair of orthogonal Helmholtz coils that were controlled using a two-channel motor controller and a laptop running custom Python scripts. A magnetometer was used monitor the magnetic field near the test sample (Infineon TLV493D-A1B6, 1000 Hz sample rate). Microswimmer responses to these field inputs were observed using an inverted, brightfield microscope and camera (PCO Panda, 25x magnification, 20 FPS, 1152×1224 pixel resolution) (Fig. \ref{fig:Coils}). During testing, the oscillating magnetic field had a measured frequency of 1.02 Hz and the rotating field had a rotational frequency of 0.25 Hz. 

 Each field input will cause the magnetic microspheres to rotate so that their internal poles align with the net magnetic field vector (Fig.~\ref{fig:Coils}b). In the case of the microswimmers, the magnetic microsphere pulls the non-magnetic microsphere as it rotates due to the DNA linkage, leading to an apparent reorientation of the microswimmer. Based on the type of field input, we expected different behaviors from the microswimmers. Specifically, we expected that an oscillating magnetic field should induce non-reciprocal gaits that result in microswimmer translation. However, the same field should not result in net translation for the symmetric magnetic microspheres.

\subsection{Fiducial Recording and Tracking}
The movement of microswimmers, non-magnetic fiducial microspheres, and ferromagnetic fiducial microspheres were tracked using TEMA (Image Systems) motion tracking software for unmarked particles. The tracker recorded the translation of the center of the fiducial microspheres and microswimmer component spheres. Of the two components tracked for the microswimmer, the center of the large ferromagnetic microsphere was used to track microswimmer displacements. The rotation of each microswimmer was also recorded to determine its rotational response to the three magnetic field inputs. Only microswimmers and fiducials that remained within the field of view for the entire experiment were included in the dataset. The net translation (distance traveled from initial position) of the non-magnetic fiducial, magnetic fiducial and microswimmer are recorded as $\Delta_{PS}$, $\Delta_{mag}$ and $\Delta_{swim}$, respectively. $\Delta x$ and $\Delta y$ represent the directional components of the particles.

We assumed that the fiducial microspheres would accurately capture any flow experienced by the microswimmers because they had the same size, density, and functional coatings as the microswimmer. Because we allowed the system to achieve steady state, we also assumed that flow between the SPT and 1xTAE layers would be approximately laminar and that thermal fluctuations would be minimal. Finally, we assumed that, even if fluid circulation were present in larger system, we could approximate the flow in a given \SI{500}{\micro\meter} by \SI{530}{\micro\meter} region captured by the microscope camera as approximately linear. 

\subsection{Estimating Microswimmer and Fiducial Drift over Time}
We estimated the drift due to fluid flow by finding the median translation vector of the non-magnetic fiducials. For each non-magnetic fiducial, we first calculated $\Delta_{PS,i}$, the translation vector representing the motion of the $i$-th non-magnetic fiducial in the dataset. 
Assuming that fluid flow acts on all fiducials equally, we calculated the median of all non-magnetic fiducial displacement vectors to determine $\Delta_{PS}$, which represents the median translation of all non-magnetic fiducials and, therefore, the median fluid flow. 

 To evaluate how accurately the flow estimate, $\Delta_{PS}$, represents the translation of individual non-magnetic fiducials, $\Delta_{PS,i}$, we calculated the Root Mean Squares Error (\ref{eqn:RMSE}). Using the x-displacement error ($E_{x,i}=\Delta_{PS,x,i}-\Delta_{PS,x} $) and y-displacement error ($E_{y,i} =\Delta_{PS,y,i}-\Delta _{PS,y}$) for each non-magnetic fiducial, RMSE was calculated over the full time vector of $s$ time steps:
\begin{gather}
RMSE = \sqrt{\frac{\sum_{j=0}^s \sum_{i=0}^n((E_{x,i}^2 + E_{y,i}^2))}{n*s}} \label{eqn:RMSE}
\end{gather}

This method was also used to estimate the drift of the magnetic fiducials. Once the drift of the magnetic and non-magnetic fiducials were estimated, they were subtracted from the microswimmer translation, $\Delta_{swim,i}$, to estimate the field-driven response of the microswimmer. We quantified this response using an RMSE for ($\Delta_{swim}-\Delta_{mag}$).


\section{Results and Discussion}
\subsection{Measuring the affect of Magnetic Field Input}

\begin{figure}[t]
\centering
\includegraphics[width=8 cm]{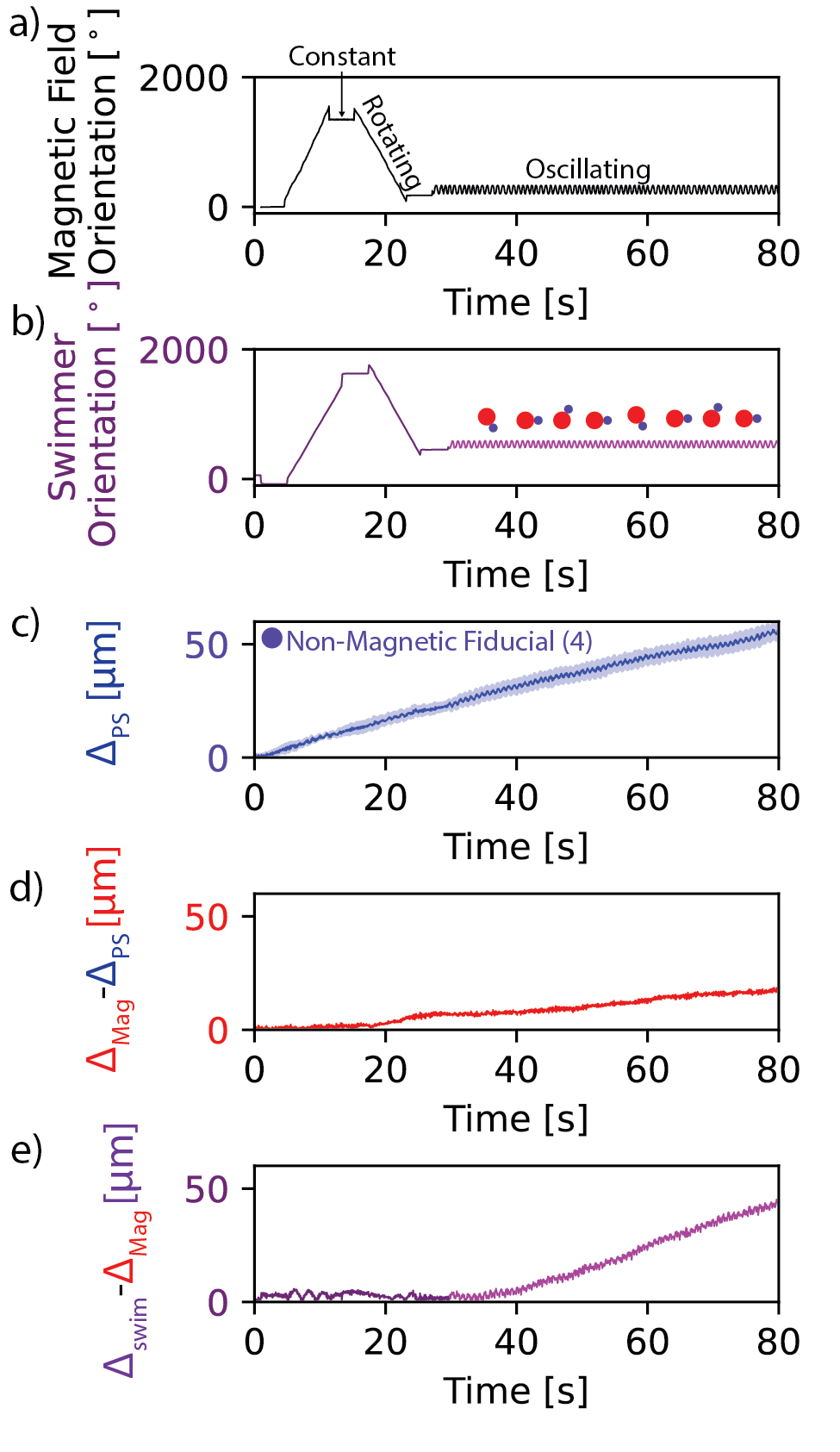}
\caption{a) The magnetic field orientation generated by the coils. b) Rotations in the magnetic field lead to rotations in the microswimmer.
c) The median and standard deviation of 4 non-magnetic fiducial displacements over time. d) The difference between the magnetic fiducial translation and the non-magnetic fiducial translation over the same time period. e) The difference between the microswimmer translation and the magnetic fiducial translation. }
\label{fig:FlowMeassure}
\end{figure}

Our ultimate goal is to create microswimmers that are capable of magnetic field-driven locomotion. By changing the external magnetic field, we align the poles of the ferromagnetic microspheres with the magnetic field vector (Fig~\ref{fig:Coils}), leading to rotation of the microswimmers and magnetic fiducials. The result of this rotational input (Fig~\ref{fig:FlowMeassure}a) is most easily seen by tracking the orientation of a microswimmer (Fig~\ref{fig:FlowMeassure}b). The magnetic fiducial will undergo the same alignment, but without a second attached body, we cannot track the orientation of this particle. 

We assessed the motion of all three particle types under constant, rotating, and oscillating magnetic field inputs (Fig~\ref{fig:FlowMeassure}b-d). 
We attributed the movement of the non-magnetic fiducials to flow that resulted from various sources in our test setup (Fig~\ref{fig:FluidWeather}b). Since these microspheres should not be affected by the applied magnetic field, we expect their motion to be relatively constant. Fig~\ref{fig:FlowMeassure}c shows the translation of four non-magnetic fiducials tracked over one experiment. 
The median translation, $\Delta_{PS}$, estimates the individual translation of the four non-magnetic fiducial motions with an RMSE of \SI{2.6}{\micro\meter} (over a range of $> $~\SI{50}{\micro\meter} traveled). It is notable that the rate of translation for the non-magnetic fiducials was relatively linear throughout the \SI{80}{\second} test, with an $R^2$ value of \SI{99.9}{\percent} for a linear fit.

Fig~\ref{fig:FlowMeassure}d shows the motion of the magnetic fiducials relative to the non-magnetic fiducials ($\Delta_{mag} - \Delta_{PS}$). Notably, the two fiducial types do not move at the same velocity. The magnetic fiducial in this experiment moved slower and in a different direction than the non-magnetic fiducials, implying that the magnetic fiducials experience an additional force most likely caused by magnetic field gradients arising from nonidealities in Helmholtz coil test setup \cite{benjaminson_buoyant_2023}.


Fig~\ref{fig:FlowMeassure}e tracks the microswimmer translation relative to the motion of the magnetic fiducials ($\Delta_{swim} - \Delta_{mag}$). We expect only the microswimmer to move differently under an oscillating field because the microswimmer is designed to break the Scallop Theorem.  Before the field starts oscillating, the microswimmer's motion approximately tracks the magnetic fiducial motion with an RMSE of \SI{2.5}{\micro\meter}. Once the the oscillatory magnetic field is introduced, the microswimmer changes its velocity and translates approximately \SI{40}{\micro\meter} relative to the magnetic fiducial. Based on this result, we conclude that the microswimmer experiences the same external forces as the magnetic fiducial during non-oscillatory magnetic fields. Therefore, this process allows us to separate microswimmer motion due to swimming from microswimmer motion due to flow and magnetic field gradients. 

\begin{figure}[t]
\centering
\includegraphics[width=8 cm]{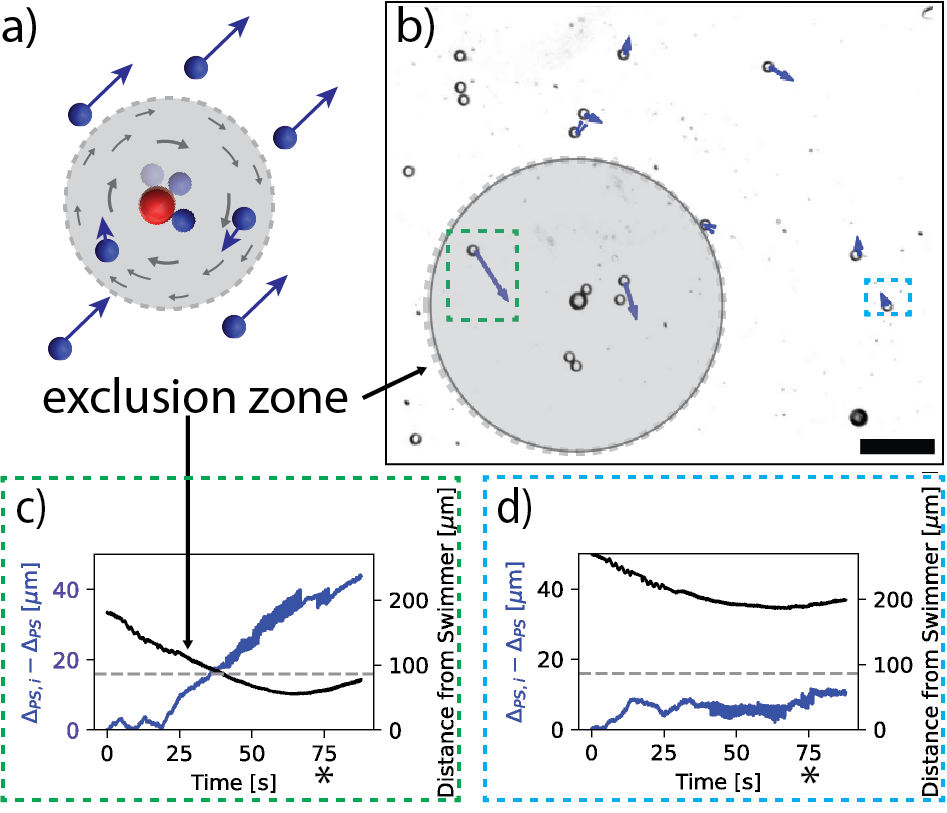}
\caption{a) When microswimmers rotate, they disturb the local flow field, affecting the motion of fiducials within an exclusion zone. b) Arrows show the error between the true translation and the estimated trajectory of each fiducial \SI{75}{\second} into the experiment. In c) the trajectory error $\Delta_{PS,i}-\Delta_{PS}$ increases as a fiducial in the exclusion zone (green box) moves closer to the microswimmer. d) Similar data for a fiducial that remains outside the exclusion zone (light blue box). Scale bar is \SI{50}{\micro\meter}.}
\label{fig:TooClose}
\end{figure}

\subsection{Local Flow Effect of Microswimmer Actuation}
Our goal is to eventually control the locomotion of colloidal microswimmers, which necessitates measuring and accounting for flow within their environment. From Fig.~\ref{fig:FlowMeassure}c we saw that the non-magnetic fiducials move consistently. This consistency allows us to assume that the net forces on all particles of a particular type are similar and to consider how those forces might affect the microswimmers. 

The only exception we observed to this assumption occurs when a fiducial microsphere comes too close to a microswimmer. In past work \cite{khaderi_magnetically-actuated_2011, diller_rotating_2011, lisa_biswal_micromixing_2004, xie_programmable_2019, zhou_ye_rotating_2014}, microswimmers were shown to create or alter flow fields within their local environment. Given these studies, it is not surprising that our microswimmers appear to affect the fluid in their immediate vicinity when they rotate or oscillate. For this reason, we included an ``exclusion zone" in our analysis. The boundary of this exclusion zone (Fig.~\ref{fig:TooClose}a) represents the area beyond which the effects of the microswimmer flow are negligible. We determined that while a fiducial is within the exclusion zone, its movement is affected by the microswimmer, and we do not use the motion of that particle for flow estimation. Additionally, when a particle leaves the exclusion zone, we only consider its translation from the location where it exited. 

In Fig.~\ref{fig:TooClose}, the estimated translation ($\Delta_{PS}$) of non-magnetic fiducials inside the exclusion zone was significantly less ($\alpha < 0.01$) accurate than the estimated translation of non-magnetic fiducials outside the exclusion zone (RMSE \SI{22.9}{\micro\meter} vs  \SI{7.0}{\micro\meter}).  This result justifies calculating $\Delta_{PS}$ using only fiducials located outside the exclusion zone. The radius of the exclusion zone was chosen by analyzing how the estimation error of $\Delta_{PS}$ changed based on the distance between the non-magnetic fiducial and the microswimmer (Fig.~\ref{fig:TooClose}c,d).  

The RMSE for $\Delta_{PS}$ in this experiment was higher than it was in the Fig.~\ref{fig:FlowMeassure} experiment, possibly because there are more non-magnetic fiducials (11 vs 4).  The RMSE may also be higher because the flow effect was larger in this region of the petri dish, with the maximum distance traveled by a non-magnetic fiducial in Fig.~\ref{fig:FlowMeassure} being \SI{57}{\micro\meter} while the maximum translation for the test in Fig.~\ref{fig:TooClose} is \SI{73}{\micro\meter}. 

After exclusions, the median motion of the non-magnetic fiducials gives us a method for estimating the local flow. Noteworthy here is our assumption that the flow within the field of view is uniform. While this assumption is false for the entire test bed (a cylindrical dish), the approximation seems reasonable in this magnified local area. If future studies consider a larger area of interest, PIV-related methods (e.g., \cite{gemmell2014new,chen2013high}) could be more useful.

\begin{figure*}[t]
\centering
\includegraphics[width=16 cm]{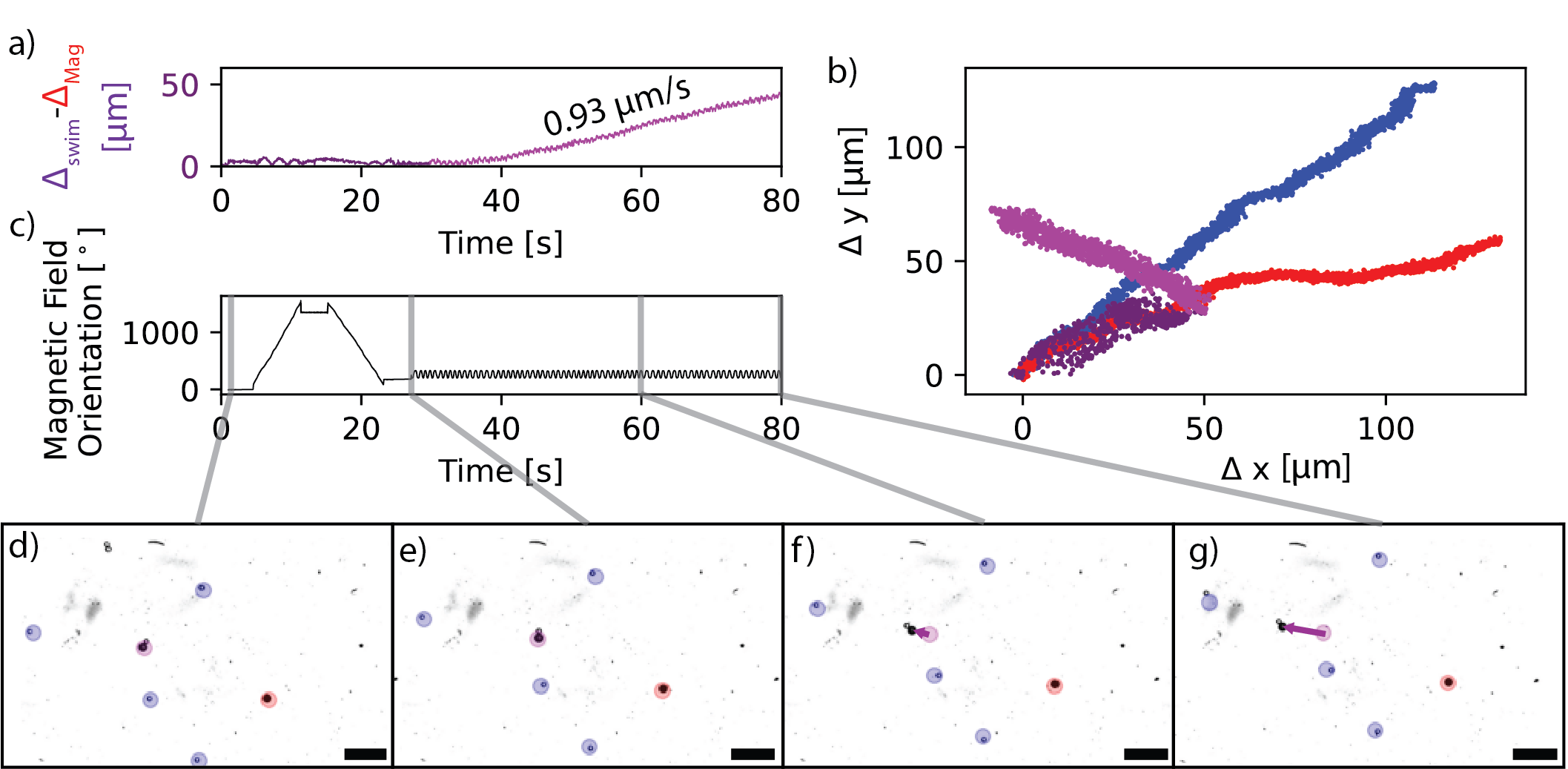}
\caption{The difference $\Delta_{swim}-\Delta_{mag}$ provides an estimate for the speed (a) and direction (b) of the microswimmer's motion. b) The direction path of all three particle types, $\Delta_{PS}$ (blue), $\Delta_{mag}$ (red) and $\Delta_{swim}$ (purple) are plotted as they are subjected to a magnetic field (c) over a 80 s test with different field types. The time during which the oscillating magnetic field occurs is highlighted in (a) and (b) using a lighter purple. (d-g) Frames from the recorded experiment, demonstrating the ability of $\Delta_{PS}$ to predict the motion of individual non-magnetic fiducials and ability of $\Delta_{mag}$ to predict the true $\Delta_{swim}$. A purple arrow is used to emphasize the  $\Delta_{swim}$-$\Delta_{mag}$ translation in (f) and (g). Scale bars are \SI{50}{\micro\meter}.}
\label{fig:SwimmerOne}
\end{figure*}

\subsection{Measuring Microswimmer Motion}

%
As seen in Fig.~\ref{fig:FlowMeassure}e, magnetic fiducials predicted microswimmer location in non-oscillatory flow fields to within \SI{2.5}{\micro\meter}. This close match between magnetic fiducial and microswimmer trajectories is also evident in Fig.~\ref{fig:SwimmerOne}b which plots the x-y trajectories of the magnetic fiducial (red) and the microswimmer (dark purple). Once the oscillatory field is introduced, the microswimmer changes its direction of travel by approximately $90^\circ$, a turn not made by the magnetic fiducial. The value of $\Delta_{swim}-\Delta_{mag}$ also increases (Fig.~\ref{fig:SwimmerOne} a,f,g). From the time the oscillating field is introduced, we estimate this microswimmer has a velocity of \SI{0.9}{\micro\meter\per\second} or approximately 0.05 BL\SI{}{\per\second}. See the supplementary video for video of this test with superimposed trajectory estimates.

\begin{figure}[t]
\centering
\includegraphics[width=0.85\linewidth]{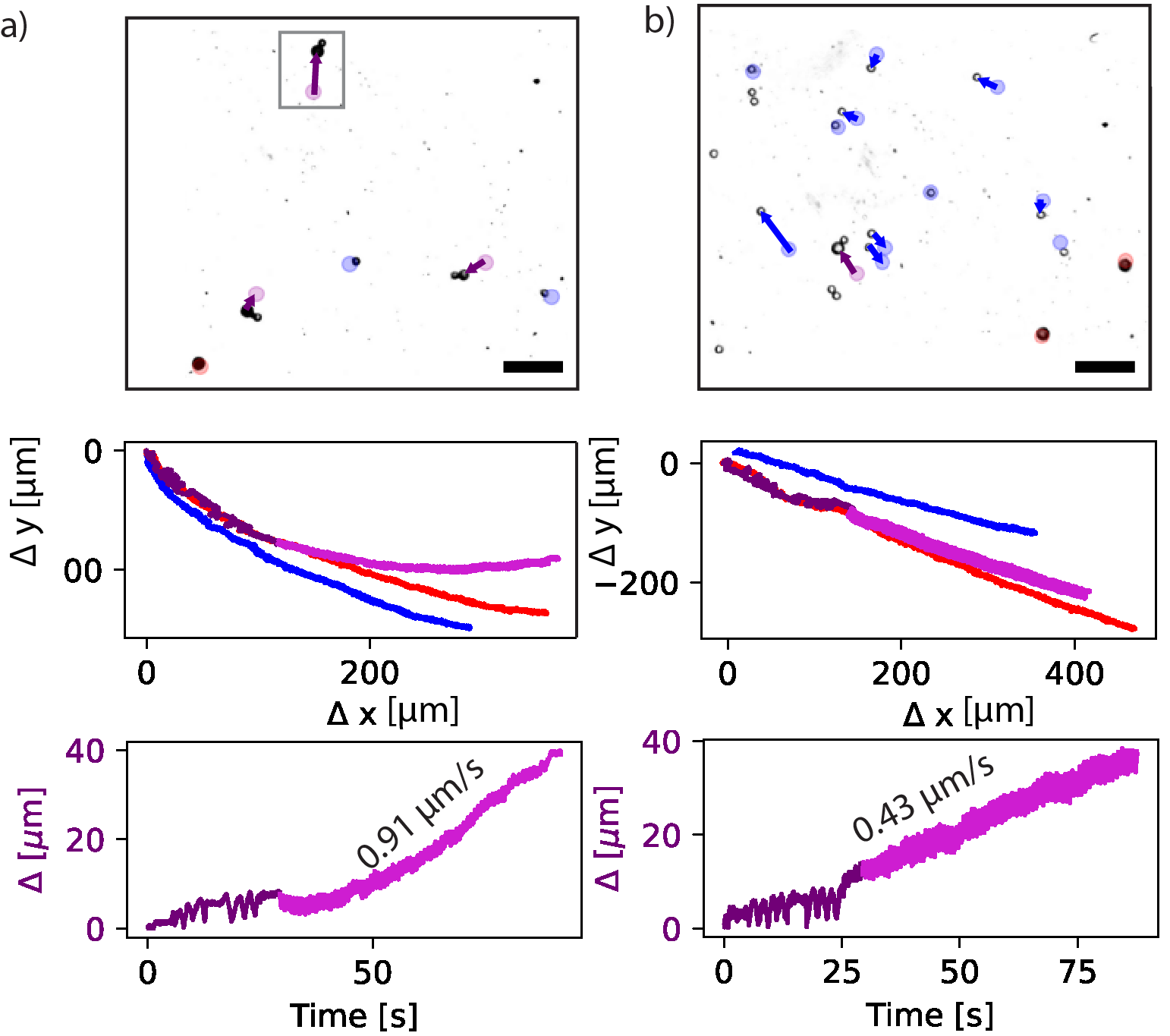}
\caption{The tests from Fig.~\ref{fig:SwimmerOne} are repeated over two more trials in (a-b). (a) includes more than one microswimmer but the microswimmer tracked is highlighted in the gray box. Scale bars are \SI{50}{\micro\meter}.}
\label{fig:SwimmerMany}
\end{figure}

Toward our goal of studying microswimmer locomotion in uncontrolled flow environments, we repeated the tests from Fig.~\ref{fig:SwimmerOne} with additional microswimmers found in different sections of the petri dish. Two example trials are shown in Fig.~\ref{fig:SwimmerMany}. Different locations led to variations in the magnetic gradient and fluid flows as highlighted in the middle row of Fig.~\ref{fig:SwimmerMany} by the different $\Delta_{PS}$ (blue) and $\Delta_{mag}$ (red) curves for each trial. Despite these flow field and gradient variations, the microswimmers all rotated to align with the field and locomoted when externally stimulated by an oscillating field.

The trial represented in Fig.~\ref{fig:SwimmerMany}b is the same trial shown in Fig.~\ref{fig:TooClose}. This trial includes multiple magnetic fiducials and shows that median motion of the magnetic fiducials predicts individual magnetic fiducial motion (with an RMSE of \SI{0.9}{\micro\meter}). For all of the microswimmers in Fig. \ref{fig:SwimmerOne} and Fig. \ref{fig:SwimmerMany} as well as three other microswimmer trials not shown here, the introduction of the oscillatory magnetic field leads to a notable increase in $\Delta_{swim}-\Delta_{mag}$. Sometimes, the microswimmer swims in a completely different direction that is easily visualized (e.g., Fig.~\ref{fig:SwimmerOne}, Fig.~\ref{fig:SwimmerMany}a). In other cases (Fig.~\ref{fig:SwimmerMany}b), the magnetic field drives the microswimmer against the fluid flow, so the translation is not as obvious visually because the microswimmer is swimming upstream. 

Microswimmers moving in different directions are observed even within the same flow field (Fig.~\ref{fig:SwimmerMany}a, top row, Supplementary Video). This is because the microswimmers align the internal pole of their magnetic microsphere to the field when exposed to constant, rotating, and oscillating magnetic field inputs (Fig.~\ref{fig:Coils}). The position of the non-magnetic microsphere relative magnetic field vector will affect the direction the microswimmer translates. The current manufacturing process does not control the direction in which the magnetic microsphere is magnetized relative to the non-magnetic microsphere, which may explain the observed variation.

Variability in manufacturing is also a contributing factor in the differences we observed in microswimmer speed. The number of DNA nanotubes that connect the microspheres is not regulated in the current fabrication process. As a result, microswimmers made within the same batch may have variable linkage stiffness, which will likely affect their swimming speeds \cite{benjaminson_buoyant_2023}. As seen in the bottom row of Fig.~\ref{fig:SwimmerMany}, the microswimmers captured in these experiments move at different speeds. The speeds shown represent a range between 0.02 BL\si{\per\second} and 0.05 BL\si{\per\second}. Future studies could explore the relationship between microswimmer magnetization and trajectory as well as the impact of DNA nanotube stiffness on microswimmer speed. In both cases, our method would facilitate a better understanding of this system by isolating the field-induced microswimmer locomotion.

\section{Conclusion}
In this work, we presented a method for measuring microswimmer locomotion within a complex flow environment using magnetic and non-magnetic fiducial microspheres. We developed an algorithm that measures fluid flow within our test environment using position data from tracked fiducials microspheres. Our algorithm then reported the net translation of microswimmers independent of flow fields or magnetic field gradients. 
To minimize error in our calculations, we estimated the size of an ``exclusion zone" in which fiducials were affected by flow from microswimmer movement, and omitted such fiducials from our calculations. We also found that other forces, which likely resulted from field gradients, contribute to differing motion between the magnetic fiducials and non-magnetic fiducials. This combined effect of the flow and the gradient force, are seen on the magnetic fiducial microspheres. In future studies of microswimmers, simultaneously tracking a magnetic fiducial and a microswimmer will allow us to isolate the field-driven effect.

Using this methodology, we demonstrated the isolation of the field-driven response on multiple microswimmers to show that they are repeatably capable of locomoting when suspended in complex flow environments. We determined that these microswimmers achieved field-driven translation only in oscillating magnetic fields, aligning with theoretical predictions for flexibly-linked microswimmers in low Reynolds environments. Our work towards isolating the field-driven response could be further explored by testing the affect of parameters such as the number of fiducials and the exclusion zone geometry on flow estimations.

This work is the first experimental study in which the locomotion of suspended colloidal microswimmers is quantified within the presence of complex fluid flow. Using this method, we can measure the ground truth movement of microswimmers, enabling future experimental investigations of structure-function relationships and path planning for colloidal microswimmers. 
This work is a first step towards creating microswimmers that are capable of accomplishing complex tasks within unstructured, real-world environments.

\bibliographystyle{IEEEtran}
\bibliography{ICRA2025_MStracking}

\end{document}